\RequirePackage{iftex}
\ifPDFTeX
\RequirePackage{cmap}
\fi
\documentclass[14pt,a4paper]{extarticle}

\ifPDFTeX
  \usepackage[T2A,T1]{fontenc}
  \usepackage[utf8]{inputenc}
\else
  \usepackage{fontspec}
\fi
\usepackage[russian,english]{babel}
\usepackage[a4paper,top=2.5cm,bottom=2.5cm,left=3.5cm,right=1.5cm]{geometry}
\usepackage{setspace}
\usepackage{indentfirst}
\usepackage{amsmath,amssymb}
\usepackage{array}
\usepackage{makecell}
\usepackage{multirow}
\usepackage{enumitem}
\usepackage{graphicx}
\usepackage{caption}
\usepackage{float}
\usepackage{textcomp}
\usepackage{xcolor}
\usepackage{hyperref}
\usepackage{tikz}
\usetikzlibrary{shapes.geometric,arrows.meta,positioning,calc,fit,backgrounds}

\hypersetup{
  colorlinks=true,
  linkcolor=black,
  citecolor=black,
  urlcolor=black
}
\addto\captionsrussian{}
\addto\captionsenglish{}

\newcolumntype{P}[1]{>{\raggedright\arraybackslash\hspace{0pt}}p{#1}}
\newcolumntype{C}[1]{>{\centering\arraybackslash\hspace{0pt}}p{#1}}
\newcommand\thinhline{\Xhline{0.1pt}}

\begin{document}
\selectlanguage{english}

{\noindent\textbf{UDC 616.5-006:004.852}}

\bigskip
\begin{center}
\textbf{Cascade Classification of Dermoscopic Images of Skin Neoplasms with Controllable Sensitivity and External Clinical Validation}
\end{center}

\bigskip
\noindent\textbf{Kozachok E.\,S.}\textsuperscript{1}, \textbf{Seregin S.\,S.}\textsuperscript{2}, \textbf{Kozachok A.\,V.}\textsuperscript{1}, \textbf{Latyshev I.\,P.}\textsuperscript{1}, \textbf{Samovarov O.\,I.}\textsuperscript{1}

\smallskip
\noindent\textsuperscript{1}Ivannikov Institute for System Programming of the Russian Academy of Sciences (ISP RAS), Moscow, Russia\\
\textsuperscript{2}Orel Oncological Dispensary, Orel, Russia

\smallskip
{\small\noindent\textit{Information about the authors:}\\
\textbf{Kozachok Elena Sergeevna}~--- specialist, ISP~RAS; ORCID: 0009-0007-9432-1663; e-mail: e.kozachok@ispras.ru \textit{(corresponding author)}.\\
\textbf{Seregin Sergey Sergeevich}~--- Cand. Sci. (Med.), oncologist, Orel Oncological Dispensary, Orel; ORCID: 0000-0002-7248-402X; e-mail: serega\_s2004@mail.ru.\\
\textbf{Kozachok Aleksandr Vasilyevich}~--- Dr. Sci. (Eng.), Professor, Head of Laboratory, ISP~RAS; ORCID: 0000-0002-6501-2008; AuthorID: 732431; Scopus ID: 57195216999; e-mail: a.kozachok@ispras.ru.\\
\textbf{Latyshev Ilya Petrovich}~--- Researcher, ISP~RAS; ORCID: 0009-0004-2485-485X; e-mail: i.latyshev@ispras.ru.\\
\textbf{Samovarov Oleg Ilgisovich}~--- Cand. Sci. (Eng.), Scientific Secretary, ISP~RAS; ORCID: 0000-0002-7006-7193; AuthorID: 17230; Scopus ID: 6508083834; e-mail: samov@ispras.ru.}

\bigskip
\noindent{\itshape \textbf{Abstract.} \\ \textbf{Purpose of research.} To compare deep learning architectures and classification schemes for automatic analysis of dermoscopic images of skin neoplasms and to assess their generalization when transferred from open international datasets to independent clinical datasets of Russian practice.

\textbf{Methods.} Four architectures~--- Vision Transformer (ViT-B/16), Swin-S, ConvNeXt-S and EfficientNetV2-S~--- were compared in three classification schemes: binary (malignant / benign), single-stage four-class (benign, melanoma~--- MEL, squamous cell carcinoma~--- SCC, basal cell carcinoma~--- BCC), and a~two-stage cascade (stage~1~--- binary triage, stage~2~--- three-class differentiation of the malignant type MEL\,/\,SCC\,/\,BCC). All models were initialised with ImageNet-pretrained weights (transfer learning with full fine-tuning) and trained on an~aggregation of open ISIC Archive datasets with a~single augmentation protocol, then evaluated on an~internal held-out sample and two independent clinical datasets (the Melanoscope AI mobile system and a~dataset from Sechenov University).

\textbf{Results.} On the internal held-out sample the binary stage attains ROC-AUC 0.952--0.966 for all architectures; when transferred to the Sechenov University clinical dataset, ROC-AUC drops to 0.797--0.893, sensitivity to 0.53--0.67, and the expected calibration error rises from 0.02 to 0.27--0.39 with underestimation of malignancy, quantifying the generalization gap in both ranking and calibration. Paired significance tests confirm a~single inter-architecture result on clinical data~--- the deficit of ViT-B/16 at the binary triage stage ($p<0.05$); at the three-class malignant-type differentiation stage no architecture has a statistically proven advantage. The cascade scheme numerically increases macro~F1 over single-stage four-class classification for most architectures, but the gain is statistically significant only for ViT-B/16~--- by recovering malignant lesions that the single-stage model assigns to the dominant benign class. On the external ISIC MILK10k benchmark, direct 11-class classification yields a~mean-class sensitivity of 0.525, confirming the limitation of the single-stage multiclass scheme for rare clinically prioritized classes.

\textbf{Conclusion.} Cascade decomposition of the pipeline with a~tunable triage threshold provides control over the sensitivity of the end-to-end decision-support system that is not attainable in standard single-stage classification assigning the image to the most probable class without additional thresholding logic, and better reproduces the clinical logic of differential diagnosis of skin neoplasms. The persistent generalization gap on external clinical datasets~--- in both ranking and calibration~--- mandates external clinical validation and recalibration prior to deployment.}


\bigskip
\section*{Introduction}\label{sec:intro}

Deep-learning-based automated analysis of dermoscopic images has reached specialist-level diagnostic accuracy on a~number of tasks~\cite{esteva2017,brinker2019,maron2019}. International open datasets~--- primarily HAM10000 and the ISIC Archive datasets~\cite{tschandl2018ham,codella2019isic}~--- have become the standard training base for most published systems. At the same time, when models trained on these datasets are transferred to clinical practice, a~drop in metrics is observed, caused by differences between the distributions of the training and test data~\cite{kouw2018introduction,daneshjou2022,combalia2022}.

From the standpoint of building a~medical information system, the design of an~ML pipeline for dermoscopy raises two independent engineering questions. The \emph{first} is the choice of architecture. In recent years the range of architectures has expanded substantially: convolutional networks (ConvNeXt, EfficientNet) have been joined by transformer architectures (ViT, Swin) with a~fundamentally different image-processing mechanism. Their comparative effectiveness on a~single dataset under identical training conditions is insufficiently studied, and published results are often incomparable because of differences in data and protocols. The \emph{second} is the choice of classification scheme. Standard single-stage multiclass classification optimizes the overall accuracy across all classes and, under strong imbalance, systematically sacrifices sensitivity to rare but clinically prioritized classes (melanoma, squamous cell carcinoma). In clinical practice the physician's decision is hierarchical: first~--- ``is there a~risk of a~malignant neoplasm?'', and only then~--- refinement of the nosological type~\cite{rotemberg2021patient}. The two-stage cascade scheme reproduces this logic.

\textbf{The aim of this work} is to compare four deep-learning architectures (ViT-B/16, Swin-S, ConvNeXt-S, EfficientNetV2-S) and three classification schemes (binary, single-stage four-class, and cascade) for dermoscopic images and to assess their generalization when transferred from open datasets to independent clinical datasets of Russian practice.

\textbf{The scientific novelty} of the work:
\begin{enumerate}
\item It is shown that cascade decomposition of the pipeline (binary triage with a~tunable threshold $\rightarrow$ nosological differentiation) provides control over the sensitivity of the end-to-end system that is not attainable in standard single-stage classification by the maximum-posterior-probability rule (assigning the image to the most probable class) without additional thresholding logic, and statistically significantly (paired bootstrap and McNemar test) improves classification consistency for the architecture least robust at the binary boundary, recovering malignant lesions absorbed by the dominant benign class.
\item The generalization gap on transfer to independent clinical datasets of Russian practice is quantitatively characterized in two dimensions~--- ranking ability (ROC-AUC) and probability calibration~--- separating the component correctable by recalibration (prior-distribution shift) from the irreducible one (feature shift).
\item A comparison of four modern architectures (two transformer and two convolutional) under identical conditions is performed with paired significance tests (DeLong, McNemar, bootstrap with multiplicity correction); it is established that on clinical data only the deficit of ViT-B/16 at the binary triage stage is reliably confirmed, whereas fine ranking of architectures at the differentiation stage is not statistically substantiated~--- which is itself a~result that limits the field's typical conclusions about the ``best'' architecture.
\end{enumerate}

\textbf{Relation to the authors' previous publications.} The methodology for forming the clinically verified dataset used in this work as the clinical test set is described in~\cite{kozachok2026dataset}; the earlier dataset with annotation of clinically significant features~--- in~\cite{kozachok2025dataset}. The mobile-dermoscopy screening methodology in which the two-stage cascade classification was first applied to the authors' own dataset and the ISIC-2019 repository is described in~\cite{kozachok2025screening}. In contrast to~\cite{kozachok2025screening}, where the cascade is used as a~ready screening tool, the present work for the first time carries out a~systematic comparison of four architectures under identical conditions and contrasts the cascade scheme with the single-stage one on identical test sets. Clinical validation of the decision-support system based on the considered pipeline is presented in a~separate publication by the authors~\cite{kozachok2026sppvr}.

\emph{Modern architectures.} The pioneering work of Esteva et al.~\cite{esteva2017} showed that a~convolutional neural network (CNN) attains dermatologist-level diagnostic accuracy in skin-neoplasm classification. Subsequent studies confirmed the consistent superiority of convolutional architectures over specialists in binary and multiclass classification~\cite{brinker2019,maron2019}; the ISIC benchmarks~\cite{codella2019isic,combalia2022} became standard venues for method comparison.

Vision Transformer (ViT)~\cite{dosovitskiy2021vit} and Swin-S~\cite{liu2021swin} differ from CNNs in the mechanism of global self-attention. This property is clinically relevant for dermoscopy: diagnostically significant patterns (atypical pigment network, blue-white veil, vascular structures) are often distributed across the whole image and require integration of global context. ConvNeXt~\cite{liu2022convnext} is a~modernized CNN whose training structure is brought closer to transformers; EfficientNetV2-S~\cite{tan2021effv2} is a~scalable CNN with progressive training. Dermoscopy-specific transformers are also being developed: DermViT~\cite{dermvit2025} adapts the ViT architecture to dermoscopic images and reports an~advantage over standard ViT and convolutional networks (accuracy $\approx$92.5\,\%, ROC-AUC $\approx$0.98 in a~four-class task). However, such results are obtained mostly on internal (in-domain) test sets; their robustness on transfer to independent clinical datasets is, as a~rule, not assessed, which is one of the emphases of the present work. A~direct comparison of the four listed architectures under identical conditions with a~generalization assessment is one of the objectives of this work.

\emph{Generalization and cascade schemes.} A~drop in quality when a~model is transferred to new clinical conditions is a~well-known generalization problem of ML systems in medicine~\cite{kouw2018introduction,daneshjou2022}. Its quantitative assessment requires validation on independent datasets different from the training distribution; in this work two clinical datasets of Russian practice are used for this purpose.

A~cascade scheme~--- a~hierarchical splitting of the task into stages with different objective functions~--- reproduces the clinical logic of the diagnostic process~\cite{rotemberg2021patient}. A~conceptually close hierarchical approach is proposed in HPDT (Hierarchical Prototypical Decision Tree)~\cite{hpdt2025}: diagnosis is organized as a~hierarchy of prototypical decisions, and on the large clinical Molemap dataset (235k images, 65~classes) the advantage of hierarchical decomposition over flat multiclass classification is shown. In contrast to such works, which are oriented toward maximizing accuracy on a~large internal dataset, the present work makes control of triage-stage sensitivity via a~tunable threshold and the assessment of generalization on transfer to independent clinical datasets the load-bearing elements. In single-stage multiclass classification, optimization of mean-class metrics under strong imbalance systematically reduces sensitivity to clinically prioritized classes; the MILK10k benchmark results (Section~\ref{sec:milk10k}) quantitatively confirm this regularity.

\section{Materials and methods}\label{sec:methods}

\subsection{Datasets}\label{sec:data}

\textbf{Training set.} The models were trained on an~aggregation of nine open ISIC Archive sub-datasets: BCN20000, Derm12345, Derm7pt, HIBA, BALD, HAM10000, ISIC~2016--2020, MILK10k, and a~group of other datasets. The split was performed at the image level in a~70\,/\,10\,/\,20 ratio (training\,/\,validation\,/\,internal held-out control); for the malignant classes, only verified cases were selected into validation and test. The resulting sizes and class distribution are given in Table~\ref{tab:split}. The sample is substantially imbalanced: benign lesions make up $\approx$81\,\% (benign\,/\,malignant ratio $\approx$4.5:1), and among the malignant nosologies the rarest class is squamous cell carcinoma ($\approx$10\,\% of malignant). This imbalance explains the gap between the high accuracy and the more modest macro-F1 values on the multiclass tasks and was compensated by class weighting during training.

\begin{table}[H]\footnotesize
\caption{Training-data sizes and class distribution (aggregated ISIC Archive sub-datasets)}\label{tab:split}
\centering
\setlength{\tabcolsep}{4pt}
\renewcommand{\arraystretch}{1.2}
\begin{tabular}{|P{40mm}|C{22mm}|C{20mm}|C{24mm}|C{20mm}|}
\hline
Class & Training & Validation & Internal held-out & Total\\
\hline
Benign & 71\,780 & 8\,324 & 16\,352 & 96\,456\\
\thinhline
Melanoma (MEL) & 7\,902 & 884 & 1\,953 & 10\,739\\
\thinhline
Squamous cell carcinoma (SCC) & 1\,691 & 192 & 347 & 2\,230\\
\thinhline
Basal cell carcinoma (BCC) & 6\,569 & 668 & 1\,401 & 8\,638\\
\hline
\textbf{Total} & \textbf{87\,942} & \textbf{10\,068} & \textbf{20\,053} & \textbf{118\,063}\\
\hline
\end{tabular}
\end{table}

The internal held-out sample (column of Table~\ref{tab:split}) serves as a~quality control in a~distribution close to the training one; in the binary task the malignant classes are merged.

\textbf{Clinical test sets.} To assess generalization, two independent clinical datasets not overlapping with the training data were used:
\begin{itemize}[leftmargin=*,topsep=2pt,itemsep=2pt]
\item \emph{Melanoscope}~--- a~clinically verified dataset obtained with the Melanoscope AI mobile dermoscopy system (the rights holder of the software complex is ISP~RAS, certificate of state registration of a~computer program No.~2026664695). The full dataset contains 1\,026 images from 443 patients, of which 39 are malignant (MEL~--- 18, BCC~--- 15, SCC~--- 6, all histologically verified); the methodology of its formation and composition are described in detail in~\cite{kozachok2026dataset} and are not repeated here. For the ML evaluation, a~subsample of 472~images was used (450~benign, 22~malignant: MEL~--- 14, SCC~--- 1, BCC~--- 7) that passed quality control and were suitable for binary and cascade evaluation.
\item \emph{Sechenov University}~--- an~independent external validation dataset formed at Sechenov First Moscow State Medical University and used in this work solely for testing: 77~images (20~benign, 57~malignant: MEL~--- 28, SCC~--- 6, BCC~--- 23). All malignant lesions are histologically verified; the annotation and confirmation of diagnoses were performed by university specialists. The dataset was not published in open repositories and does not overlap with the training data. It is enriched with malignant lesions (the oncological profile of referrals), which makes it a~strict test of sensitivity.
\end{itemize}

\textbf{Justification of the evaluation subsample.} The use of a~subsample of the Melanoscope dataset for the ML evaluation is a~deliberate quality-control procedure, not a~selection by result. The criteria for forming the subsample are set independently of the model and rely on the dataset's service fields~\cite{kozachok2026dataset}: images with a~low \texttt{image\_quality\_score} (capture artifacts, defocus, foreign objects in the frame) and records that did not reach the required \texttt{verification\_stage} were excluded. The subsample preserves all three nosological classes of malignant lesions (MEL, SCC, BCC) and the benign\,/\,malignant proportion characteristic of the outpatient flow, and therefore remains representative of the target scenario. The metrics in all tables are computed strictly on this fixed subsample; its size is stated explicitly. The small size of the malignant part of the clinical datasets (especially SCC) determines the wide confidence intervals of the metrics and is taken into account when interpreting the results (Section~\ref{sec:lim}).

\subsection{Classification schemes}\label{sec:schemes}

\begin{figure}[H]
\centering
\resizebox{\linewidth}{!}{%
\begin{tikzpicture}[
  font=\small,
  box/.style={rectangle,rounded corners=2pt,draw,align=center,minimum height=10mm,inner sep=4pt},
  dec/.style={diamond,aspect=2,draw,align=center,inner sep=1pt},
  >={Stealth[length=2.2mm]}
]
\node[box,fill=blue!6] (img) {Dermoscopic\\image};
\node[box,fill=green!8,right=12mm of img] (s1) {Stage~1\\binary\\classifier};
\node[dec,fill=yellow!12,right=12mm of s1] (d) {$P(\text{mal.})$\\$\geq\theta$?};
\node[box,fill=green!8,right=13mm of d] (s2) {Stage~2\\3-class\\MEL\,/\,SCC\,/\,BCC};
\node[box,fill=blue!6,below=10mm of d] (ben) {Benign};
\node[box,right=12mm of s2] (out) {Malignancy type};
\draw[->] (img) -- (s1);
\draw[->] (s1) -- (d);
\draw[->] (d) -- node[above]{yes} (s2);
\draw[->] (d) -- node[right]{no} (ben);
\draw[->] (s2) -- (out);
\end{tikzpicture}}
\caption{Two-stage cascade classification scheme. Stage~1 triages by $P(\text{malignancy})$ with a~tunable threshold~$\theta$; stage~2 is activated at $P(\text{malignancy})\geq\theta$ and outputs the nosological class MEL\,/\,SCC\,/\,BCC. Stage~2 in the standalone (3-class) scheme is applied only to malignant lesions}\label{fig:cascade}
\end{figure}
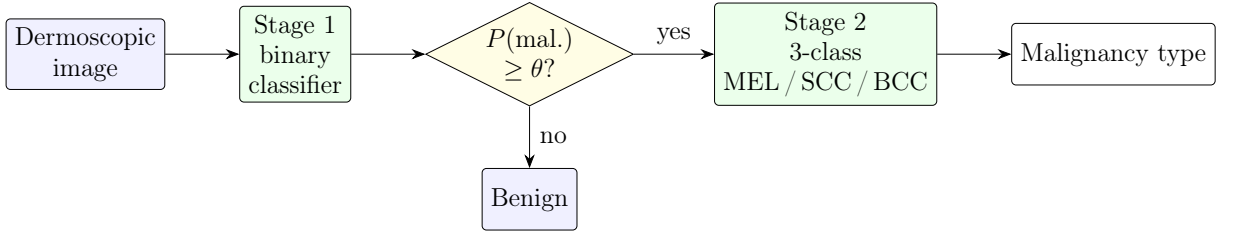

Three schemes were compared (Figure~\ref{fig:cascade}):
\begin{itemize}[leftmargin=*,topsep=2pt,itemsep=2pt]
\item \textbf{Binary} (2~classes): malignant / benign. Used both as a~standalone scheme and as stage~1 of the cascade.
\item \textbf{Single-stage four-class} (4~classes): benign, MEL, SCC, BCC~--- a~single classifier without splitting into stages (the baseline comparison scheme).
\item \textbf{Two-stage cascade} (the proposed pipeline): stage~1~--- binary triage by the malignancy probability $P(\text{mal.})$ with a~tunable threshold~$\theta$ (default value $\theta=0.50$); images for which $P(\text{mal.})\geq\theta$ are routed to stage~2~--- three-class differentiation MEL\,/\,SCC\,/\,BCC trained on malignant lesions. Benign lesions filtered out by stage~1 are output directly. As stage~1, in all cascade variants a~\emph{single fixed} binary classifier based on ConvNeXt-S was used~--- the most robust binary model on the internal sample (Table~\ref{tab:stage1}); only the stage-2 differentiator varied. This choice isolates the contribution of the nosological differentiator and provides all cascade variants with the same strong input triage stage.
\end{itemize}

Separating the objective functions between stages makes it possible to tune the sensitivity threshold of stage~1 independently of the trade-offs of joint multiclass optimization. In a~screening scenario the threshold can be lowered (e.g., to $\theta=0.15$) to increase sensitivity; this principle is consistent with the mobile-dermoscopy screening methodology~\cite{kozachok2025screening} and is used in the three-zone patient-routing algorithm in the clinical validation of the system~\cite{kozachok2026sppvr}.

\subsection{Compared architectures and training protocol}\label{sec:archs}

Four architectures were compared (Table~\ref{tab:archs}), initialized with ImageNet-pretrained weights.

\begin{table}[H]\footnotesize
\caption{Compared architectures}\label{tab:archs}
\centering
\setlength{\tabcolsep}{3pt}
\renewcommand{\arraystretch}{1.2}
\begin{tabular}{|P{28mm}|P{34mm}|C{17mm}|P{40mm}|}
\hline
Architecture & Backbone (\texttt{torchvision}) & Type & Key feature\\
\hline
ViT-B/16~\cite{dosovitskiy2021vit} & \texttt{vit\_b\_16} & Transformer & Global self-attention; 16$\times$16 patch tokens\\
\thinhline
Swin-S~\cite{liu2021swin} & \texttt{swin\_s} & Transformer & Hierarchical attention in shifted windows\\
\thinhline
ConvNeXt-S~\cite{liu2022convnext} & \texttt{convnext\_small} & Convolutional & Modernized CNN, aligned with transformers\\
\thinhline
EfficientNetV2-S~\cite{tan2021effv2} & \texttt{efficientnet\_v2\_s} & Convolutional & Scalable CNN with progressive training\\
\hline
\end{tabular}
\end{table}

\textbf{Transfer learning.} All four architectures were initialized with ImageNet-pretrained weights (transfer learning): the convolutional or transformer backbone is transferred from the general image-recognition task, while the classification head for the target number of classes is initialized randomly and trained from scratch. The backbone is not frozen~--- all layers of the network are fine-tuned (full fine-tuning), which allows low-level features to be adapted to the specifics of dermoscopic images. Training from scratch (without pretraining) on the available amount of data is not competitive for such large models, so transferring ImageNet features is a~standard and obligatory element of the methodology.

\begin{figure}[H]
\centering
\includegraphics[width=0.7\linewidth]{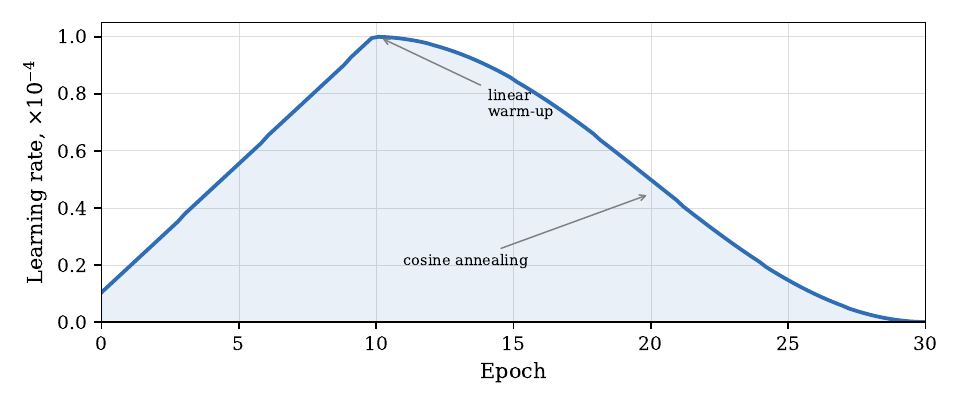}
\caption{Learning-rate schedule: linear warm-up followed by cosine annealing (AdamW, peak value $10^{-4}$)}\label{fig:lr}
\end{figure}

\textbf{Training protocol} (identical for all architectures and schemes). A~unified head was placed on top of each backbone: \texttt{Linear}($d_\text{features}\!\to\!512$)\,$\to$\,\texttt{BatchNorm1d}(512)\,$\to$\,\texttt{Linear}(512$\to$\,num\_classes), dropout~$=0$. The optimizer is AdamW ($\text{lr}=10^{-4}$, weight decay~$=10^{-2}$); the learning-rate schedule is a~linear warm-up (10\,000~steps, \texttt{SequentialLR}) followed by cosine annealing (Figure~\ref{fig:lr}); 30~epochs, batch size~16, mixed precision \texttt{bf16}, fixed random seed~$=42$, a~single NVIDIA~A100 GPU; the final model is the best checkpoint by the validation metric \texttt{val/f1\_macro}. The loss function is cross-entropy; the image size is $224\times224$~pixels; normalization by the training-set statistics; a~single set of augmentations (geometric~--- transpose, flips, rotations; photometric~--- ColorJitter, PlanckianJitter, RandomGamma, CLAHE; noise and blur~--- GaussNoise, GaussianBlur, MedianBlur; CoarseDropout). The class imbalance was compensated by weighting. Training and testing were carried out in 2026 on ISP~RAS resources.

\textbf{Metrics and statistical analysis.} ROC-AUC, sensitivity, specificity, F-measure~--- for the binary stage; per-class and macro ROC-AUC, accuracy, macro-F1~--- for the multiclass schemes. Owing to the small size of the malignant part of the clinical datasets, exact 95\,\% Clopper--Pearson confidence intervals are given for sensitivity and specificity. For a~formal comparison of architectures, the probabilistic predictions of each model for each image were stored, which made it possible to perform paired significance tests on identical samples. The difference in binary-stage ROC-AUC between architecture pairs was assessed with the DeLong test (two-sided; the 95\,\% AUC confidence intervals are also built on it); the difference in accuracy~--- with the exact McNemar test on paired predictions. For the stage-2 multiclass metrics (macro-F1, macro ROC-AUC), to which the DeLong test is inapplicable, the significance of differences and the 95\,\% confidence intervals were obtained by a~paired stratified bootstrap (bootstrap~--- a~resampling method by drawing with replacement; 5000~iterations, resampling preserving class sizes, common indices for all models). Additionally, the calibration of the binary-stage probabilities was assessed~--- the Brier score and the expected calibration error (ECE). The significance level is $\alpha=0.05$; for multiple pairwise comparisons, Holm-corrected $p$-values are additionally reported.

\begin{figure}[H]
\centering
\includegraphics[width=\linewidth]{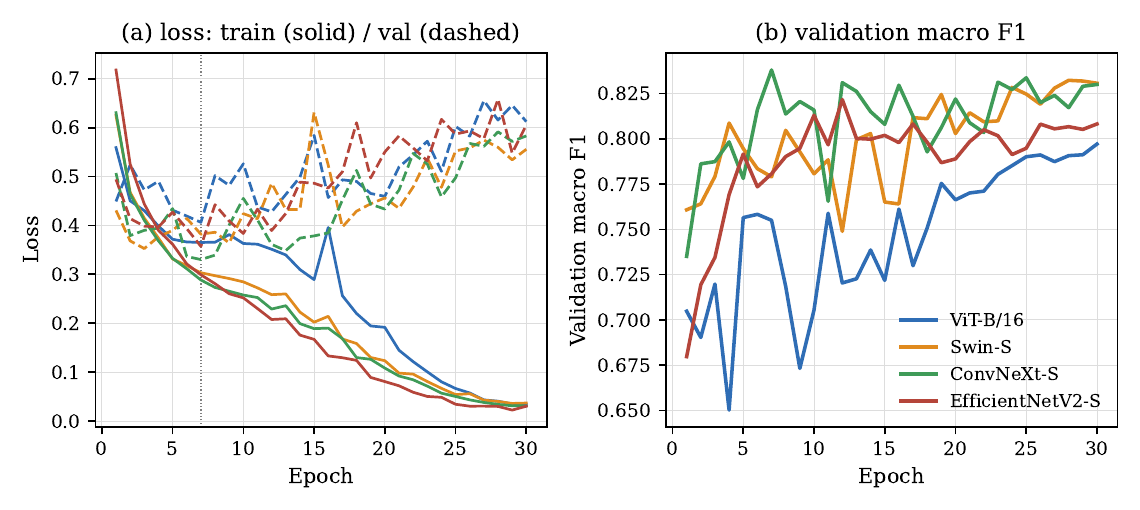}
\caption{Learning curves (stage~2, MEL\,/\,SCC\,/\,BCC): (a)~training (solid) and validation (dashed) loss; (b)~validation macro~F1. The vertical dotted line marks the onset of overfitting ($\approx$epoch~7)}\label{fig:train}
\end{figure}

\textbf{Training dynamics.} The training curves for the four architectures (stage~2, three-class task) are shown in Figure~\ref{fig:train}. Owing to initialization with ImageNet weights, the validation macro-F1 reaches 0.70--0.81 already within the first 3--5~epochs~--- direct evidence of the effectiveness of feature transfer. After $\approx$the 7th epoch the validation loss begins to grow while the training loss continues to decrease (the classical overfitting regime), so the final model is selected by the best validation checkpoint rather than by the last epoch. The transformer and convolutional architectures show similar dynamics; Swin-S and ConvNeXt-S reach a~somewhat higher validation plateau.

\section{Results}\label{sec:results}

\subsection{Stage~1: binary classification}\label{sec:stage1}

\begin{figure}[H]
\centering
\includegraphics[width=\linewidth]{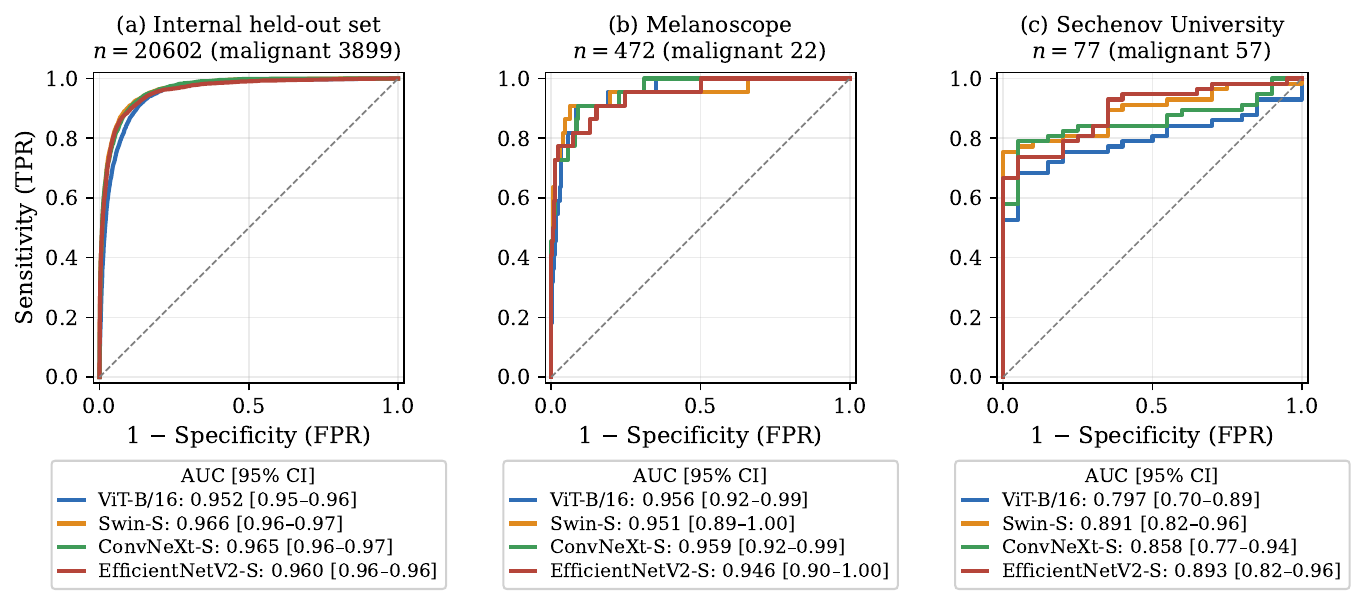}
\caption{Binary-stage ROC curves (malignant / benign) for four architectures on three test sets; the legend shows the area under the curve with its 95\,\% DeLong confidence interval. The generalization gap on the Sechenov University dataset is visible: the curves drop and the confidence intervals widen as the sample size decreases}\label{fig:stage1roc}
\end{figure}

The binary-stage metrics (malignant / benign) are given in Table~\ref{tab:stage1}; the ROC curves with the areas under them and 95\,\% confidence intervals (DeLong) for all architectures and datasets are shown in Figure~\ref{fig:stage1roc}.

\begin{table}[H]\footnotesize
\caption{Binary classification metrics (malignant / benign) by architecture and test set}\label{tab:stage1}
\centering
\setlength{\tabcolsep}{4pt}
\renewcommand{\arraystretch}{1.2}
\begin{tabular}{|P{36mm}|C{16mm}|C{20mm}|C{20mm}|C{14mm}|}
\hline
Architecture & AUC & Sens. & Spec. & F1\\
\hline
\multicolumn{5}{|l|}{\textit{Internal held-out set}}\\
\hline
ViT-B/16 & 0.952 & 0.710 & 0.955 & 0.746\\
\thinhline
Swin-S & \textbf{0.966} & \textbf{0.802} & 0.956 & \textbf{0.805}\\
\thinhline
ConvNeXt-S & 0.965 & 0.782 & 0.955 & 0.792\\
\thinhline
EfficientNetV2-S & 0.960 & 0.789 & 0.958 & 0.801\\
\hline
\multicolumn{5}{|l|}{\textit{Melanoscope}}\\
\hline
ViT-B/16 & 0.956 & 0.591 & 0.971 & 0.542\\
\thinhline
Swin-S & 0.951 & 0.727 & 0.973 & 0.640\\
\thinhline
ConvNeXt-S & \textbf{0.959} & 0.727 & \textbf{0.987} & \textbf{0.727}\\
\thinhline
EfficientNetV2-S & 0.946 & \textbf{0.773} & 0.973 & 0.667\\
\hline
\multicolumn{5}{|l|}{\textit{Sechenov University}}\\
\hline
ViT-B/16 & 0.797 & 0.526 & \textbf{1.000} & 0.690\\
\thinhline
Swin-S & 0.891 & 0.614 & \textbf{1.000} & 0.761\\
\thinhline
ConvNeXt-S & 0.858 & \textbf{0.667} & 0.950 & \textbf{0.792}\\
\thinhline
EfficientNetV2-S & \textbf{0.893} & 0.561 & \textbf{1.000} & 0.719\\
\hline
\end{tabular}
\end{table}

On the internal held-out sample all four architectures show similar quality (ROC-AUC 0.952--0.966), with a~slight advantage of Swin-S. On the Melanoscope dataset the quality is preserved (ROC-AUC 0.946--0.959), but on the Sechenov University dataset a~noticeable decline is observed: ROC-AUC drops to 0.797--0.893, and sensitivity to 0.53--0.67. This is direct quantitative evidence of the generalization gap between the training distribution and independent clinical practice. Notably, on the most difficult external dataset the convolutional and window-transformer architectures (ConvNeXt-S, EfficientNetV2-S, Swin-S) outperform ViT-B/16 in ROC-AUC, i.e., at the binary triage stage the advantage of global self-attention does not manifest.

Owing to the small size of the malignant part of the clinical datasets, Table~\ref{tab:ci} gives exact 95\,\% (Clopper--Pearson) confidence intervals for the sensitivity and specificity of the binary stage. The sensitivity confidence intervals are wide (the malignant sample sizes are 22 and 57, respectively), so the inter-architecture differences in sensitivity overlap; what is robust at the level of individual metrics is the systemic conclusion~--- the decline in sensitivity on both clinical datasets compared with the internal sample. Specificity is estimated on a~large number of benign lesions and has narrow intervals. A~formal significance test of the differences in ROC-AUC and accuracy is given below.

\begin{table}[H]\footnotesize
\caption{Binary-stage sensitivity and specificity on the clinical datasets with exact 95\,\% Clopper--Pearson confidence intervals}\label{tab:ci}
\centering
\setlength{\tabcolsep}{3pt}
\renewcommand{\arraystretch}{1.2}
\begin{tabular}{|P{30mm}|C{36mm}|C{36mm}|}
\hline
Architecture & Sens. [95\,\% CI] & Spec. [95\,\% CI]\\
\hline
\multicolumn{3}{|l|}{\textit{Melanoscope (malignant $n=22$, benign $n=450$)}}\\
\hline
ViT-B/16 & 0.591 [0.364--0.793] & 0.971 [0.951--0.985]\\
\thinhline
Swin-S & 0.727 [0.498--0.893] & 0.973 [0.954--0.986]\\
\thinhline
ConvNeXt-S & 0.727 [0.498--0.893] & 0.987 [0.971--0.995]\\
\thinhline
EfficientNetV2-S & 0.773 [0.546--0.922] & 0.973 [0.954--0.986]\\
\hline
\multicolumn{3}{|l|}{\textit{Sechenov University (malignant $n=57$, benign $n=20$)}}\\
\hline
ViT-B/16 & 0.526 [0.390--0.660] & 1.000 [0.832--1.000]\\
\thinhline
Swin-S & 0.614 [0.476--0.740] & 1.000 [0.832--1.000]\\
\thinhline
ConvNeXt-S & 0.667 [0.529--0.786] & 0.950 [0.751--0.999]\\
\thinhline
EfficientNetV2-S & 0.561 [0.424--0.693] & 1.000 [0.832--1.000]\\
\hline
\end{tabular}
\end{table}

\textbf{Statistical significance of differences between architectures.} To separate robust differences from sample noise, a~pairwise comparison of architectures was performed with the DeLong test (ROC-AUC) and the exact McNemar test (accuracy) on identical samples. The results for the most difficult external dataset (Sechenov University) are given in Table~\ref{tab:delong}. On this dataset ViT-B/16 is statistically significantly inferior in ROC-AUC to two architectures~--- Swin-S ($\Delta\text{AUC}=0.094$, $p=0.006$) and EfficientNetV2-S ($\Delta\text{AUC}=0.096$, $p=0.035$); the gap from ConvNeXt-S has the same direction but does not reach significance ($p=0.087$). At the same time, the differences \emph{within} the top three (Swin-S, ConvNeXt-S, EfficientNetV2-S) are not statistically confirmed: in particular, the formally noticeable superiority of Swin-S~0.891 over ConvNeXt-S~0.858 is not significant ($\Delta\text{AUC}=0.033$, $p=0.15$) and should be treated as sample noise. Thus, what is statistically proven is precisely the \emph{systemic} conclusion~--- at the binary triage stage on a~difficult external dataset ViT-B/16 is inferior to the convolutional and window-transformer architectures, whereas the ranking within the latter is not substantiated. On the Melanoscope dataset (only 22~malignant lesions) none of the pairwise ROC-AUC differences are significant (all $p>0.3$). On the internal held-out sample ($n=20{,}602$) all inter-architecture differences are significant owing to the large size, but are small in magnitude ($\Delta\text{AUC}\leq0.013$) and of no practical importance.

\begin{table}[H]\footnotesize
\caption{Pairwise architecture comparison at the binary stage, Sechenov University dataset ($n=77$): $\Delta$ROC-AUC with DeLong $p$-value, and exact McNemar $p$-value for accuracy. Significant differences ($p<0.05$) are shown in bold}\label{tab:delong}
\centering
\setlength{\tabcolsep}{4pt}
\renewcommand{\arraystretch}{1.2}
\begin{tabular}{|P{50mm}|C{20mm}|C{20mm}|C{24mm}|}
\hline
Architecture pair & $\Delta$AUC & $p$ (DeLong) & $p$ (McNemar)\\
\hline
ViT-B/16 / Swin-S & $-0.094$ & \textbf{0.006} & 0.180\\
\thinhline
ViT-B/16 / ConvNeXt-S & $-0.061$ & 0.087 & \textbf{0.039}\\
\thinhline
ViT-B/16 / EfficientNetV2-S & $-0.096$ & \textbf{0.035} & 0.791\\
\thinhline
Swin-S / ConvNeXt-S & $+0.033$ & 0.154 & 0.754\\
\thinhline
Swin-S / EfficientNetV2-S & $-0.002$ & 0.952 & 0.607\\
\thinhline
ConvNeXt-S / EfficientNetV2-S & $-0.035$ & 0.292 & 0.227\\
\hline
\end{tabular}
\end{table}

The given $p$-values relate to individual comparisons and are not corrected for multiplicity. After the Holm correction over six pairs, the only difference that remains significant is for the pair ViT-B/16~/~Swin-S (DeLong, corrected $p=0.036$); the borderline results (DeLong ViT-B/16~/~EfficientNetV2-S and McNemar ViT-B/16~/~ConvNeXt-S, raw $p\approx0.04$) do not retain significance after correction. This is consistent with a~conservative reading: the statistically most robust effect is precisely the deficit of ViT-B/16 relative to Swin-S at the binary triage stage.

\subsection{Calibration of binary-stage probabilities}\label{sec:calibration}

\begin{figure}[H]
\centering
\includegraphics[width=\linewidth]{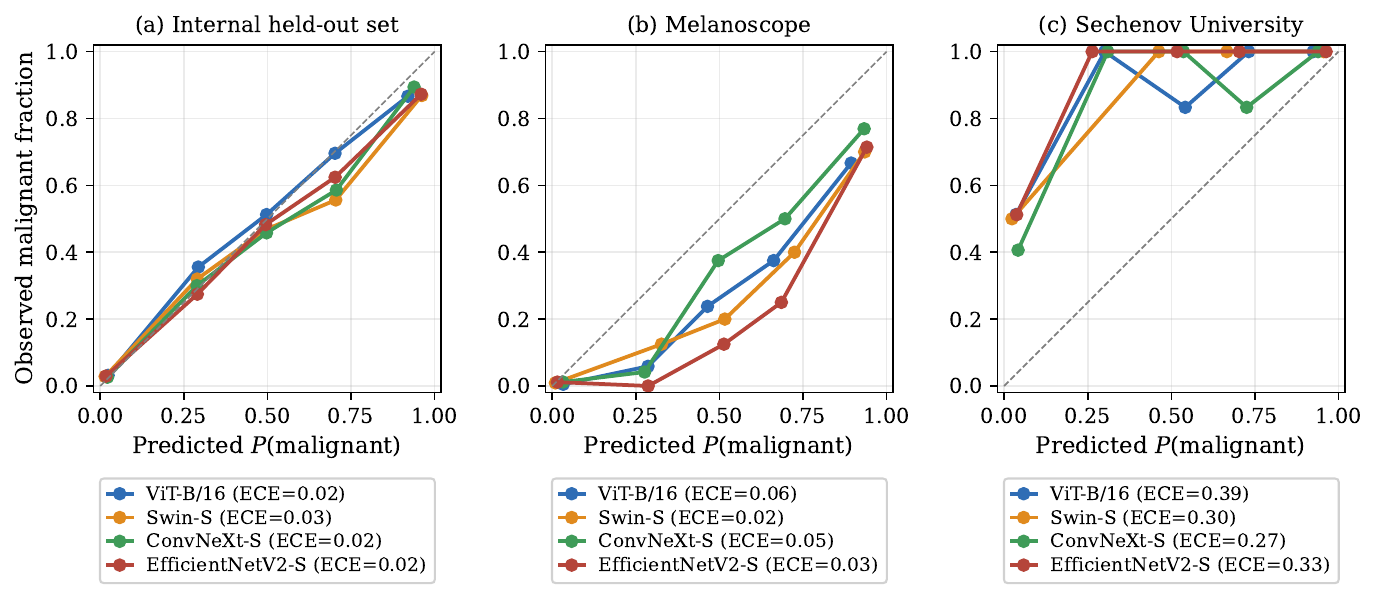}
\caption{Binary-stage reliability diagrams: observed malignant fraction vs predicted $P(\text{malignancy})$ (grouped into 5~bins for display; the ECE values in the legend and in Table~\ref{tab:calib} use 10~bins). Perfect calibration is the diagonal. On the Sechenov University dataset the curves lie above the diagonal (malignancy is underestimated)}\label{fig:calib}
\end{figure}

ROC-AUC characterizes only the \emph{ranking} ability of a~model and is invariant to a~monotone transformation of probabilities. However, the cascade scheme makes its decision by the absolute value of $P(\text{mal.})$ relative to the threshold~$\theta$, so \emph{calibration}~--- the correspondence of predicted probabilities to actual frequencies~--- is of practical importance. Calibration was assessed by the Brier score and the expected calibration error (ECE, expected calibration error; the deviation of the mean predicted probability from the actual fraction, averaged over 10~bins); the results are given in Table~\ref{tab:calib}, the reliability diagrams~--- in Figure~\ref{fig:calib}.

\begin{table}[H]\footnotesize
\caption{Binary-stage probability calibration: Brier score and expected calibration error (ECE). Lower is better}\label{tab:calib}
\centering
\setlength{\tabcolsep}{4pt}
\renewcommand{\arraystretch}{1.2}
\begin{tabular}{|P{36mm}|C{16mm}|C{16mm}|C{16mm}|C{16mm}|C{16mm}|C{16mm}|}
\hline
\multirow{2}{*}{Architecture} & \multicolumn{2}{c|}{Internal} & \multicolumn{2}{c|}{Melanoscope} & \multicolumn{2}{c|}{Sechenov Univ.}\\
\cline{2-7}
 & Brier & ECE & Brier & ECE & Brier & ECE\\
\hline
ViT-B/16 & 0.067 & 0.016 & 0.038 & 0.062 & 0.311 & 0.386\\
\thinhline
Swin-S & 0.057 & 0.030 & 0.027 & 0.022 & 0.258 & 0.296\\
\thinhline
ConvNeXt-S & 0.056 & 0.015 & 0.027 & 0.046 & \textbf{0.216} & \textbf{0.271}\\
\thinhline
EfficientNetV2-S & 0.057 & 0.025 & 0.030 & 0.028 & 0.285 & 0.333\\
\hline
\end{tabular}
\end{table}

On the internal sample and the Melanoscope dataset the models are well calibrated (ECE 0.015--0.062). On the Sechenov University dataset calibration sharply degrades: ECE rises to 0.27--0.39, the Brier score~--- to 0.22--0.31. The reliability diagrams (Figure~\ref{fig:calib}) show that on this dataset the curves lie \emph{above} the diagonal: at a~low predicted probability of malignancy the actual fraction of malignant lesions turns out to be substantially higher~--- the models systematically \emph{underestimate} malignancy. The calibration error has a~clinically unfavorable direction: at a~fixed threshold~$\theta$ some truly malignant lesions receive $P(\text{mal.})<\theta$ and are filtered out, which is directly related to the observed drop in sensitivity (Section~\ref{sec:stage1}).

The source of this error should be separated. A~significant part of it is a~shift of the prior class distribution (in the Sechenov University dataset 74\,\% of lesions are malignant versus $\approx$21\,\% in training), correctable by base-rate correction: correcting the logits for the difference in class prior probabilities reduces the ECE roughly by half (for ViT-B/16 from 0.386 to 0.184, for ConvNeXt-S from 0.271 to 0.102). This means that part of the calibration gap is a~consequence of the composition of the clinical sample rather than a~feature-distribution shift, and is removable by tuning the threshold and the class prior probabilities on the target data. However, even after correcting the prior distribution a~residual calibration error remains (0.10--0.19), and single-parameter temperature scaling hardly removes it; in parallel, the drop in ranking ability persists (ROC-AUC from 0.95--0.97 to 0.80--0.89), invariant to any monotone recalibration. Thus, the generalization gap is composed of a~correctable component (prior-distribution shift) and a~component irreducible by recalibration (residual calibration error and ROC-AUC decline); the practical conclusion is that before clinical deployment both recalibration on the target distribution and separate tuning of the stage-1 threshold (Section~\ref{sec:threshold}) are mandatory, but they remove the gap only partially.

\subsection{Stage~2: three-class differentiation of malignant lesions}\label{sec:stage2}

\begin{figure}[H]
\centering
\includegraphics[width=0.6\linewidth]{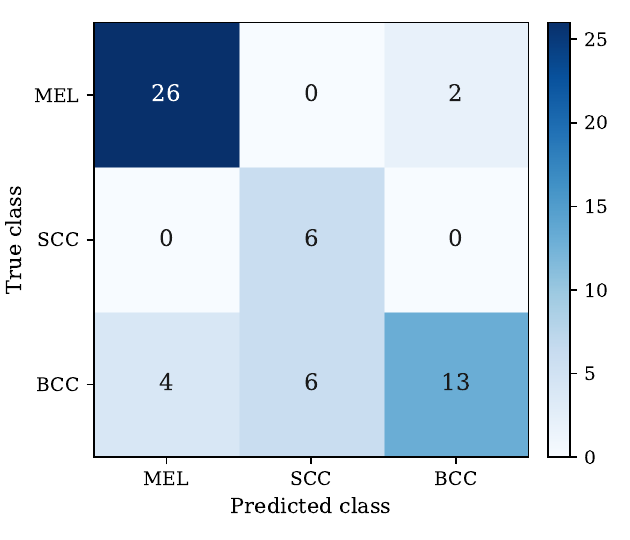}
\caption{Confusion matrix of ViT-B/16 in three-class differentiation (MEL\,/\,SCC\,/\,BCC) on the Sechenov University dataset ($n=57$)}\label{fig:cm2}
\end{figure}

The metrics of three-class differentiation (MEL\,/\,SCC\,/\,BCC) on malignant lesions are given in Table~\ref{tab:stage2}. The confusion matrix of ViT-B/16 on the Sechenov University dataset is shown in Figure~\ref{fig:cm2}.

\begin{table}[H]\footnotesize
\caption{Malignant-type differentiation (MEL\,/\,SCC\,/\,BCC): macro ROC-AUC, accuracy, macro~F1}\label{tab:stage2}
\centering
\setlength{\tabcolsep}{4pt}
\renewcommand{\arraystretch}{1.2}
\begin{tabular}{|P{36mm}|C{22mm}|C{22mm}|C{22mm}|}
\hline
Architecture & macro AUC & Accuracy & macro F1\\
\hline
\multicolumn{4}{|l|}{\textit{Internal held-out set}}\\
\hline
ViT-B/16 & 0.949 & 0.860 & 0.794\\
\thinhline
Swin-S & 0.963 & 0.890 & 0.841\\
\thinhline
ConvNeXt-S & \textbf{0.966} & \textbf{0.895} & \textbf{0.845}\\
\thinhline
EfficientNetV2-S & 0.953 & 0.873 & 0.816\\
\hline
\multicolumn{4}{|l|}{\textit{Melanoscope, $n=22$}}\\
\hline
ViT-B/16 & \textbf{1.000} & \textbf{0.955} & \textbf{0.965}\\
\thinhline
Swin-S & 0.849 & 0.818 & 0.574\\
\thinhline
ConvNeXt-S & 0.835 & 0.773 & 0.564\\
\thinhline
EfficientNetV2-S & 0.943 & 0.773 & 0.552\\
\hline
\multicolumn{4}{|l|}{\textit{Sechenov University, $n=57$}}\\
\hline
ViT-B/16 & 0.944 & 0.790 & 0.749\\
\thinhline
Swin-S & \textbf{0.952} & \textbf{0.877} & \textbf{0.823}\\
\thinhline
ConvNeXt-S & 0.941 & 0.790 & 0.744\\
\thinhline
EfficientNetV2-S & 0.928 & 0.790 & 0.740\\
\hline
\end{tabular}
\end{table}

In nosological-type differentiation the picture differs from the binary stage. On the Melanoscope dataset ViT-B/16 formally shows the highest consistency (macro-F1~$=$~0.965 versus 0.55--0.57 for the other architectures), but this value cannot be regarded as a~reliable estimate: the dataset contains only $n=22$ malignant lesions, of which a~single SCC case, so macro-averaging over three classes is extremely unstable (one correct SCC prediction gives a~perfect contribution of that class). On the larger-by-malignant Sechenov University dataset ($n=57$) Swin-S numerically leads (macro-F1 0.823 [95\,\% bootstrap CI 0.704--0.930] versus 0.749 [0.643--0.855] for ViT-B/16), but the paired bootstrap does not confirm the significance of the difference ($\Delta\text{F1}=0.074$, $p=0.16$); none of the pairwise differences in macro-F1 or macro ROC-AUC between the four architectures on this dataset reaches $p<0.05$. Consequently, on clinical data no architecture has a~statistically proven advantage at the nosological-type differentiation stage. The main source of errors is the erroneous assignment of basal cell carcinoma to melanoma and squamous cell carcinoma (Figure~\ref{fig:cm2}): of 23~BCC cases, 13 are correctly classified, whereas melanoma and squamous cell carcinoma are recognized almost without error.

The per-class stage-2 metrics on the Sechenov University dataset are given in Table~\ref{tab:stage2pc}. Melanoma and squamous cell carcinoma are recognized with high sensitivity (recall 0.86--1.00) for all architectures; the bottleneck is basal cell carcinoma (recall 0.52--0.78), which is consistent with the confusion matrix (Figure~\ref{fig:cm2}): part of BCC is erroneously assigned to melanoma and squamous cell carcinoma. The low precision for the SCC class is due to the small number of cases of this class ($n=6$).

\begin{table}[H]\footnotesize
\caption{Per-class stage-2 metrics (recall\,/\,precision) on the Sechenov University dataset. Class sizes: MEL~--- 28, SCC~--- 6, BCC~--- 23}\label{tab:stage2pc}
\centering
\setlength{\tabcolsep}{3pt}
\renewcommand{\arraystretch}{1.2}
\begin{tabular}{|P{30mm}|C{16mm}|C{16mm}|C{16mm}|C{16mm}|C{16mm}|C{16mm}|}
\hline
\multirow{2}{*}{Architecture} & \multicolumn{2}{c|}{MEL} & \multicolumn{2}{c|}{SCC} & \multicolumn{2}{c|}{BCC}\\
\cline{2-7}
 & recall & prec. & recall & prec. & recall & prec.\\
\hline
ViT-B/16 & 0.93 & 0.87 & 1.00 & 0.50 & 0.57 & 0.87\\
\thinhline
Swin-S & 0.96 & 0.96 & 0.83 & 0.56 & \textbf{0.78} & 0.90\\
\thinhline
ConvNeXt-S & 0.96 & 0.84 & 1.00 & 0.50 & 0.52 & 0.92\\
\thinhline
EfficientNetV2-S & 0.86 & 0.96 & 0.83 & 0.45 & 0.70 & 0.76\\
\hline
\end{tabular}
\end{table}

\subsection{Cascade scheme versus single-stage classification}\label{sec:compare_modes}

\begin{figure}[H]
\centering
\includegraphics[width=0.96\linewidth]{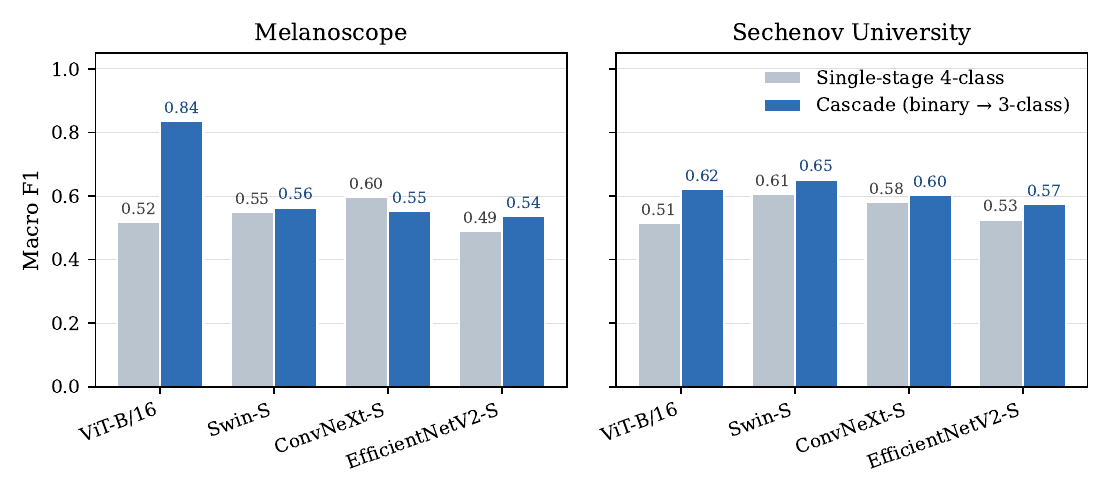}
\caption{Comparison of macro~F1 of single-stage four-class and cascade (binary~$\rightarrow$~3-class) schemes across four architectures on two clinical datasets}\label{fig:compare}
\end{figure}

The end-to-end comparison of the cascade scheme (binary triage $\rightarrow$ three-class differentiation) with single-stage four-class classification by macro-F1 on the clinical datasets is given in Table~\ref{tab:compare} and Figure~\ref{fig:compare}.

\begin{table}[H]\footnotesize
\caption{Cascade (binary $\rightarrow$ 3-class) vs.\ single-stage 4-class classification: macro~F1 on the clinical datasets}\label{tab:compare}
\centering
\setlength{\tabcolsep}{4pt}
\renewcommand{\arraystretch}{1.2}
\begin{tabular}{|P{34mm}|C{24mm}|C{22mm}|C{24mm}|C{22mm}|}
\hline
\multirow{2}{*}{Architecture} & \multicolumn{2}{c|}{Melanoscope} & \multicolumn{2}{c|}{Sechenov Univ.}\\
\cline{2-5}
 & single-stage & cascade & single-stage & cascade\\
\hline
ViT-B/16 & 0.517 & \textbf{0.836} & 0.514 & \textbf{0.622}\\
\thinhline
Swin-S & 0.550 & 0.564 & 0.607 & \textbf{0.652}\\
\thinhline
ConvNeXt-S & \textbf{0.598} & 0.552 & 0.581 & \textbf{0.603}\\
\thinhline
EfficientNetV2-S & 0.491 & \textbf{0.538} & 0.526 & \textbf{0.575}\\
\hline
\end{tabular}
\end{table}

The cascade scheme increases macro-F1 for most architecture\,$\times$\,dataset combinations; the most pronounced gain is for ViT-B/16 (on the Melanoscope dataset 0.517~$\rightarrow$~0.836). The exception is ConvNeXt-S on the Melanoscope dataset, where the single-stage scheme is slightly higher (0.598 versus 0.552). The advantage of the cascade is explained by the fact that stage~2 is trained to differentiate only the malignant types and is not forced to simultaneously hold the boundary with the numerous benign class.

A~paired significance test of the gain (macro-F1~--- stratified bootstrap; accuracy~--- exact McNemar test on paired correctness on identical images) shows that the cascade gain is statistically significant only for ViT-B/16. Robust is the result on the Sechenov University dataset ($\Delta\text{F1}=+0.107$; bootstrap $p=0.024$; McNemar $p=0.039$; the malignant class sizes are sufficient for the estimate). On the Melanoscope dataset formally $\Delta\text{F1}=+0.319$, but this value is inflated by the degenerate structure of the subsample: about $0.25$ of the macro-F1 gain is contributed by a~single SCC case ($n=1$), which the single-stage scheme assigns to benign and the cascade~--- correctly; therefore the magnitude of the gain on Melanoscope should not be interpreted quantitatively. For Swin-S, ConvNeXt-S, and EfficientNetV2-S the cascade gain is small ($|\Delta\text{F1}|\leq0.05$) and does not reach significance on any dataset ($p\geq0.2$).

The mechanism of the gain on the Sechenov University dataset can be traced through the end-to-end confusion matrices. In the single-stage regime ViT-B/16 erroneously assigns to benign 30 of 57~malignant lesions~--- the largest fraction among the architectures (Swin-S~--- 22, ConvNeXt-S~--- 23, EfficientNetV2-S~--- 29), which is consistent with its least robust binary boundary (Section~\ref{sec:stage1}). In the cascade the triage task is solved not by the architecture itself but by the single fixed binary ConvNeXt-S (stage~1, Section~\ref{sec:schemes}), which at $\theta=0.50$ passes 38 of 57~malignant lesions to differentiation (erroneously filters out 19); on the input it has filtered, the ViT-B/16 differentiator correctly recognizes 10 melanomas of~28 (versus 6 in the single-stage regime) and 12 basal cell carcinomas of~23 (versus 8). Thus, the cascade gain for ViT-B/16 is due not to an~improvement in fine differentiation (at stage~2 this architecture does not outperform the others, Section~\ref{sec:stage2}), but to the replacement of its own weak binary separation by a~common strong input triage stage that recovers the malignant lesions that the single-stage model ``absorbs'' into the dominant benign class. Since stage~1 is the same in all cascade variants, the gain is naturally concentrated in the architecture least robust at the binary boundary in the single-stage regime (ViT-B/16); for architectures already strong in binary separation, the common input triage stage yields no significant gain.

\begin{figure}[H]
\centering
\includegraphics[width=0.62\linewidth]{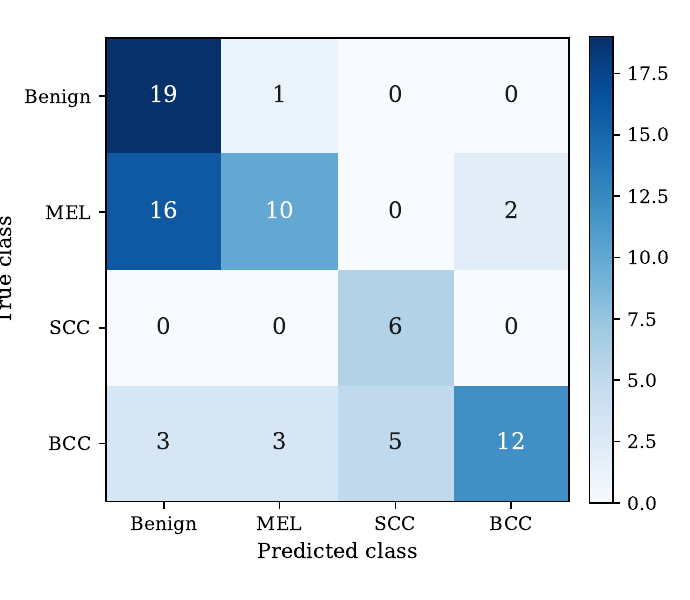}
\caption{End-to-end cascade confusion matrix, Sechenov University dataset ($\theta=0.50$). Stage~1 is a~fixed ConvNeXt-S binary classifier (the ``Benign'' column is its direct output), stage~2 is the ViT-B/16 differentiator (columns MEL\,/\,SCC\,/\,BCC). Rows~--- true class, columns~--- final system decision}\label{fig:cascadecm}
\end{figure}

The end-to-end operation of the cascade is illustrated by the combined $4\times4$ confusion matrix (Figure~\ref{fig:cascadecm}): the rows are the true class (benign, MEL, SCC, BCC), the columns are the final system decision (benign~--- the direct output of stage~1; MEL\,/\,SCC\,/\,BCC~--- the stage-2 class). It is seen that the main loss of sensitivity occurs at stage~1: part of the melanomas and basal cell carcinomas ($P(\text{mal.})<\theta$) is filtered into benign before differentiation; among the lesions that passed the threshold, the nosological type is determined correctly in most cases. This confirms that to increase the sensitivity of the end-to-end system the primary lever is the stage-1 threshold rather than the stage-2 accuracy (Section~\ref{sec:threshold}).

\subsection{Tuning the stage-1 operating point}\label{sec:threshold}

The tunable threshold~$\theta$ of stage~1 (the fixed binary ConvNeXt-S) makes it possible to control the sensitivity/specificity trade-off of the end-to-end system. On the Sechenov University dataset, lowering the threshold from the neutral $\theta=0.50$ to $\theta=0.15$ raises the sensitivity of malignant-lesion triage from 0.667 to 0.789 at the cost of a~specificity decrease from 0.950 to 0.900; on the Melanoscope dataset~--- from 0.727 to 0.909 with a~specificity decrease from 0.987 to 0.898. This confirms that separate tuning of the stage-1 threshold is a~standard mechanism for adapting the system to the required safety level (priority of sensitivity in the screening regime), not attainable in a~standard single-stage multiclass model that assigns the image to the most probable class without additional thresholding logic. The sensitivity values given in Table~\ref{tab:stage1} correspond to the neutral operating point $\theta=0.50$; lowering the threshold is consistent with the results of the clinical validation, where at a~lowered routing threshold no false-negative cases for malignant neoplasms were recorded~\cite{kozachok2026sppvr}.

\subsection{External benchmark: direct 11-class classification on ISIC MILK10k}\label{sec:milk10k}

For an~independent assessment of the quality ceiling of direct multiclass classification, the results of the ISP~RAS team (rank~27) on the open ISIC MILK10k benchmark~\cite{isic2024milk10k} (EfficientNetV2-S architecture) are considered. The evaluation was performed by the competition server on the closed (hidden) test part of MILK10k, which is not included in the publicly available part of the dataset; therefore, although the publicly available MILK10k images are present in the training set (Section~\ref{sec:data}), the overlap of the training data with the benchmark's test part is excluded by the design of the competition itself. The full results are given in Table~\ref{tab:milk10k}.

\begin{table}[H]\scriptsize
\caption{11-class EfficientNetV2-S on the ISIC MILK10k benchmark (Ivannikov ISP~RAS, rank~27)}\label{tab:milk10k}
\centering
\setlength{\tabcolsep}{1.5pt}
\renewcommand{\arraystretch}{1.2}
\resizebox{\linewidth}{!}{%
\begin{tabular}{|P{26mm}|C{11mm}|C{11mm}|C{11mm}|C{11mm}|C{11mm}|C{11mm}|C{11mm}|C{11mm}|C{11mm}|C{11mm}|C{11mm}|C{11mm}|}
\hline
Metric & Mean & AKIEC & BCC & BEN & BKL & DF & INF & MAL & MEL & NV & SCCKA & VASC\\
\hline
AUC & 0.860 & 0.867 & 0.957 & 0.745 & 0.818 & 0.959 & 0.858 & 0.603 & 0.933 & 0.935 & 0.912 & 0.875\\
\thinhline
Accuracy & 0.934 & 0.896 & 0.887 & 0.987 & 0.839 & 0.975 & 0.977 & 0.979 & 0.946 & 0.950 & 0.856 & 0.983\\
\thinhline
Sensitivity & 0.525 & 0.606 & 0.934 & 0.143 & 0.310 & 0.667 & 0.273 & 0.100 & \textbf{0.667} & 0.635 & 0.846 & 0.600\\
\thinhline
Specificity & 0.962 & 0.946 & 0.880 & 1.000 & 0.979 & 0.979 & 0.994 & 0.998 & 0.970 & 0.988 & 0.859 & 0.987\\
\thinhline
BMA & 0.527 & \multicolumn{11}{c|}{---}\\
\hline
\end{tabular}}
\end{table}

\noindent{\small\textit{Classes: AKIEC~--- actinic keratosis/intraepithelial carcinoma; BCC~--- basal cell carcinoma; BEN~--- other benign; BKL~--- benign keratosis; DF~--- dermatofibroma; INF~--- infectious; MAL~--- other malignant; MEL~--- melanoma; NV~--- nevus; SCCKA~--- squamous cell carcinoma/keratoacanthoma; VASC~--- vascular. BMA~--- balanced multiclass accuracy.}}

\medskip
In direct 11-class classification, the high overall accuracy (0.934) is combined with a~low mean-class sensitivity (0.525) and balanced multiclass accuracy (BMA~$=$~0.527): the high accuracy masks the unsatisfactory detectability of rare classes. This illustrates the general drawback of the single-stage multiclass scheme and justifies the separation of a~binary triage stage with a~tunable threshold.

\section{Discussion}\label{sec:discussion}

\subsection{The architecture depends on the stage}\label{sec:disc_arch}

The only statistically confirmed inter-architecture difference on clinical data relates to the binary triage stage: on the most difficult external dataset (Sechenov University) ViT-B/16 is significantly inferior in ROC-AUC to the window-transformer Swin-S and the convolutional EfficientNetV2-S (DeLong test, $p=0.006$ and $p=0.035$; the gap from ConvNeXt-S is of the same direction, $p=0.087$). At the same time, the differences within the top three are not significant (e.g., Swin-S~0.891 versus ConvNeXt-S~0.858, $p=0.15$), so the data do not allow ranking these three architectures relative to one another. At the nosological-type differentiation stage (MEL\,/\,SCC\,/\,BCC) on the clinical datasets, no architecture has a~statistically significant advantage in macro-F1 or macro ROC-AUC (paired bootstrap, all $p\geq0.05$); the numerically higher value of ViT-B/16 on the Melanoscope dataset is due to the degenerate structure of the subsample (a~single SCC case) and is not reproduced on the Sechenov University dataset. Thus, only the deficit of ViT-B/16 at the binary triage stage on difficult external data is reliably established~--- possibly because the ``malignant\,/\,benign'' triage relies on local features, whereas the global self-attention of ViT performs worse than hierarchical and convolutional models under a~distribution shift. This result determines the engineering decision in the proposed pipeline: at the triage stage, where the inter-architecture differences are significant, the most robust binary model (ConvNeXt-S) is fixed, whereas at the differentiation stage, where no architecture has a~proven advantage, the choice of the second stage is not critical and may be determined by other considerations (speed, model size). The thesis of separately assigning the ``optimal'' architecture to each stage should be treated as a~design principle rather than a~proven result: significance is established only for the deficit of ViT-B/16 at the binary boundary, while confirmation of fine ranking at the differentiation stage requires larger clinical samples.

\subsection{Cascade versus single-stage classification}\label{sec:disc_cascade}

The gain of the cascade scheme is architecture-specific. Although numerically macro-F1 grows for most combinations (Table~\ref{tab:compare}), the paired tests (bootstrap for macro-F1, McNemar test for accuracy) confirm the significance of the gain only for ViT-B/16 (Sechenov University dataset, $p<0.05$); for Swin-S, ConvNeXt-S, and EfficientNetV2-S the gain does not exceed sample noise. This result should not be interpreted as a~universal benefit of cascading~--- it points to a~specific mechanism. ViT-B/16 in the single-stage four-class regime systematically loses on the ``malignant\,/\,benign'' separation (Section~\ref{sec:stage1}, where it is significantly inferior in ROC-AUC) and therefore assigns to benign the largest fraction of malignant lesions among the architectures. The cascade moves the binary boundary into a~separate stage with its own decision rule, removing the competition of malignant subtypes with the numerous benign class, and thereby recovers exactly the deficit that manifests in ViT-B/16. Architectures already robust at the binary boundary do not have this structural defect, and separating an~independent triage stage yields them no significant macro-F1 gain~--- although the sensitivity-control mechanism via a~tunable threshold (Section~\ref{sec:threshold}) remains useful for all architectures regardless of the F1 gain. The MILK10k benchmark (Section~\ref{sec:milk10k}) shows the flip side of the single-stage approach~--- a~drop in mean-class sensitivity to 0.525 at a~high overall accuracy. The cascade scheme with a~tunable stage-1 threshold (Section~\ref{sec:threshold}) provides an~engineering mechanism for sensitivity control that is not directly attainable in a~standard single-stage model that selects the most probable class without additional thresholding logic.

\subsection{The generalization gap}\label{sec:disc_gap}

The decline in binary-stage ROC-AUC from 0.95--0.97 on the internal sample to 0.80--0.89 on the Sechenov University dataset and the drop in sensitivity to 0.53--0.67 is a~quantitative characterization of the generalization gap when a~model trained on open international datasets is transferred to independent clinical practice. The gap also manifests in the second dimension~--- probability calibration: the expected calibration error rises from 0.02 on the internal sample to 0.27--0.39 on the Sechenov University dataset (Section~\ref{sec:calibration}), and in a~clinically unfavorable direction~--- the models underestimate malignancy. These two dimensions have a~different nature: the ROC-AUC decline reflects a~feature-distribution shift and is invariant to recalibration, whereas the calibration gap is partly explained by a~shift of the base class rate in the clinical sample (correctable by prior-distribution correction, Section~\ref{sec:calibration}) and only in the residual part~--- by calibration error proper. The single augmentation protocol during training plays the role of regularization but does not remove the ranking gap. The obtained result is consistent with known observations of the decline of dermatological ML systems on independent clinical data~\cite{daneshjou2022,combalia2022} and underscores the obligatory nature of external clinical validation before deployment.

\subsection{Limitations}\label{sec:lim}

\emph{The main limitation is the small size of the clinical samples and the limited statistical power.} The Melanoscope subsample contains only 22~malignant lesions (including a~single SCC case), the Sechenov set~--- 57. At such sizes the per-class and macro metrics have wide confidence intervals, and individual values are unstable: the macro-F1~$=$~0.965 of ViT-B/16 on Melanoscope is technically achievable with a~single correct prediction for the SCC class ($n=1$) and is not a~reliable estimate. The paired tests performed (DeLong for ROC-AUC, exact McNemar for accuracy, stratified bootstrap for the multiclass metrics) make it possible to separate robust differences from noise, but their power is limited: at the nosological-type differentiation stage the differences between architectures do not reach significance, and it cannot be excluded that some of them are real but not detected at the current sizes (type~II error). What is confirmed is only the deficit of ViT-B/16 at the binary triage stage on the Sechenov University dataset and the systemic decline of metrics on transfer to clinical data. In view of this, the obtained estimates on the clinical datasets should be regarded as a~preliminary external validation rather than as a~sufficient basis for clinical deployment without further prospective multicenter verification.

\emph{Multiplicity of comparisons and power.} After the Holm correction over six pairwise comparisons, only the difference ViT-B/16~/~Swin-S at the binary triage stage remains robust (Section~\ref{sec:stage1}); the borderline results ($p\approx0.04$) do not retain significance. Given the small size of the clinical samples, a~type~II error for real but undetected differences cannot be excluded. Expanding the clinical samples (a~multicenter dataset) and pre-registering the set of comparisons are subjects of further work.

Other limitations. The comparison is limited to two clinical datasets~--- generalization to other institutions and regions requires multicenter verification. A~single training protocol and a~fixed set of augmentations were used; per-architecture hyperparameter tuning could change the relative order. The MILK10k benchmark was obtained on a~different class nomenclature and serves only as an~illustration of the limit of the single-stage multiclass scheme rather than a~direct comparison with the proposed pipeline.

\section*{Conclusion}\label{sec:conclusion}

This work carried out a~systematic comparison of four deep-learning architectures (ViT-B/16, Swin-S, ConvNeXt-S, EfficientNetV2-S) and three classification schemes for dermoscopic images on identical training data and identical test sets, including two independent clinical datasets of Russian practice.

\textbf{Main results:}
\begin{enumerate}
\item The relative advantage of an~architecture depends on the stage, but only the deficit of ViT-B/16 at the binary triage stage on the most difficult external dataset (Sechenov University) is statistically reliably established: it is significantly inferior in ROC-AUC to Swin-S and EfficientNetV2-S (DeLong, $p<0.05$), whereas the differences within the top three are not significant. At the nosological-type differentiation stage no architecture has a~statistically proven advantage on clinical data. Therefore, in the proposed pipeline the triage stage is fixed on the most robust binary model (ConvNeXt-S), and the choice of the differentiator architecture is not critical; the separate assignment of the ``optimal'' architecture to the stages remains a~design principle rather than a~proven result.
\item The benefit of the cascade scheme (binary triage $\rightarrow$ three-class differentiation) is architecture-specific: a~statistically significant macro-F1 gain over single-stage four-class classification is confirmed only for ViT-B/16 (Sechenov University dataset, $\Delta\text{F1}=+0.107$, $p<0.05$, bootstrap and McNemar test), whereas for Swin-S, ConvNeXt-S, and EfficientNetV2-S the gain is not statistically confirmed. The mechanism is the recovery of malignant lesions that the single-stage model assigns to the dominant benign class: the cascade is most useful to the architecture that separates malignant and benign worst of all. Regardless of the F1 gain, the cascade provides all architectures with a~sensitivity-control mechanism via the tunable stage-1 threshold.
\item The generalization gap is quantitatively characterized in two dimensions: the binary-stage ROC-AUC declines from 0.95--0.97 on the internal sample to 0.80--0.89 on the Sechenov University dataset (feature shift, irreducible by recalibration), and the calibration error (ECE) rises from 0.02 to 0.27--0.39 with underestimation of malignancy (partly explained by a~base class-rate shift and correctable by recalibration on the target distribution). This justifies the obligatory nature of external validation and recalibration before deployment.
\item The external ISIC MILK10k benchmark (rank~27) additionally illustrates the limitation of direct 11-class classification in sensitivity to rare clinically prioritized classes (mean-class sensitivity 0.525 at a~high overall accuracy).
\end{enumerate}

\textbf{Practical significance.} Comparing architectures and schemes under identical conditions provides reproducible guidelines for designing ML pipelines for skin-neoplasm diagnosis in medical information systems. The cascade scheme with separate tuning of the triage stage provides sensitivity control~--- an~obligatory condition for safe clinical use. The persistent generalization gap on external clinical datasets determines the need for clinical validation before deployment.

\section*{Ethics}\label{sec:ethics}

The study is retrospective and was performed on de-identified dermoscopic images: the data were processed in anonymized form that does not allow patient identification and were used for research purposes only. The work was conducted in accordance with the principles of the Declaration of Helsinki of the World Medical Association. The clinical-dataset images were obtained during routine diagnostic examination; no personal data were included in the analysis.

\section*{Funding}\label{sec:funding}

The work was carried out within the research plan of the Ivannikov Institute for System Programming of the Russian Academy of Sciences; the study had no external funding.

\section*{Conflict of interests}\label{sec:coi}

The authors declare no conflict of interests related to the publication of this article.

\end{document}